\documentclass{article}
\usepackage[utf8]{inputenc}
\usepackage[T1]{fontenc}
\usepackage[accepted]{sysml2019}
\usepackage{newtxtext}
\usepackage{newtxmath}
\usepackage[varqu,varl,scaled=0.95]{zi4}
\usepackage{microtype}
\usepackage{graphicx}
\usepackage{xcolor}
\usepackage{subfigure}
\usepackage{booktabs}
\usepackage{flushend}

\usepackage[hidelinks]{hyperref}

\usepackage{amsmath}
\usepackage{amssymb}
\usepackage{bbm}
\usepackage{nicefrac}
\usepackage{soul}
\usepackage{enumitem}

\newcommand{\ourtitle}{Full deep neural network training on a pruned weight budget}\sysmltitlerunning{\ourtitle}
\newcommand{\pubtag}{\vspace{1em}\emph{Appears in the Proceedings of the 2$^\mathit{nd}$ Conference on Systems and Machine Learning (SysML 2019)}}
\newcommand{\ours}{Dropback}

\begin{document}

\twocolumn[
\sysmltitle{\ourtitle}

\sysmlsetsymbol{equal}{*}

\begin{sysmlauthorlist}
\sysmlauthor{Maximilian Golub}{ubc,mb}
\sysmlauthor{Guy Lemieux}{ubc}
\sysmlauthor{Mieszko Lis}{ubc}
\end{sysmlauthorlist}

\sysmlaffiliation{ubc}{Department of Electrical Engineering, The University of British Columbia, Vancouver, Canada}
\sysmlaffiliation{mb}{Mercedes-Benz Research \& Development North America, Seattle, WA, USA}

\sysmlcorrespondingauthor{Mieszko Lis}{mieszko@ece.ubc.ca}

\sysmlkeywords{machine learning, training, pruning, compression, SysML}

\vskip 0.3in

\begin{abstract}
We introduce a DNN training technique that learns only a fraction of the full parameter set without incurring an accuracy penalty. To do this, our algorithm constrains the total number of weights updated during backpropagation to those with the highest total gradients. The remaining weights are not tracked, and their initial value is regenerated at every access to avoid storing them in memory. This can dramatically reduce the number of off-chip memory accesses during both training and inference, a key component of the energy needs of DNN accelerators. By ensuring that the total weight diffusion remains close to that of baseline unpruned SGD, networks pruned using our technique are able to retain state-of-the-art accuracy across network architectures — including networks previously identified as difficult to compress, such as Densenet and WRN. With ResNet18 on ImageNet, we observe an 11.7$\times$ weight reduction with no accuracy loss, and up to 24.4$\times$ with a small accuracy impact.
\end{abstract}
]

\printAffiliationsAndNotice{}

\section{Introduction}
On-device inference has become an important market, and a number of vendors offer low-power deep neural network accelerators that reduce computation costs to specifically target mobile and embedded applications~\citep{Movidius:2017,Samsung:2018:Exynos9,Qualcomm:2017:NPE,Kingsley:2017:A11}. To enable inference despite the comparatively smaller memories and reduced memory bandwidths in mobile devices — an order of magnitude less capacity and two orders of magnitude less bandwidth than a datacentre-class GPU — many researchers have proposed pruning, quantizing, and compressing neural networks, often trading off accuracy versus network size~\citep[etc.]{lecun_optimal_1990,hassibioptimalsurgeon1993,hanlearningweightsconnections2015,handeep2015,wuquantized2015,choitowards2016,alvarezcompression-aware2017,gecompressing2017,zhouincremental2017,luothinet:2017,EnergyAwarePruning:CVPR:2017}.

Comparatively little attention, however, has been paid to the problem of on-device \emph{training}, which has so far been limited to simple models~\citep[e.g.,][]{Apple:2017:FaceDetection}. Training takes much more energy than inference: one iteration on a single sample involves roughly thrice as many weight accesses, and typically many iterations on many samples are required. Off-chip memory accesses dominate, costing hundreds of times more energy than computation: in a 45nm process, for example, accessing a 32-bit value from DRAM costs over 700$\times$ more energy than an 32-bit floating-point compute operation~\citep[640pJ vs.~0.9pJ,][]{haneie:2016}, with an even larger gap at smaller process nodes. Existing pruning, quantization, and reduction techniques are only applied \emph{after} training, and do not ameliorate this energy bottleneck.

Most of the off-chip memory references during training come from accessing activations and weights. The ratio depends on the amount of weight reuse available in the DNN architecture being trained: for example, in an MLP weights are not reused within a single training sample and typically need more bandwidth than activations, while in a CNN each filter is reused for an entire layer and there can be an order of magnitude more activation accesses than weight accesses.

Several prior works have been able to reduce the footprint of activations by an order of magnitude, via techniques like gradient checkpointing~\citep{chentraining2016}, accelerator architectures that elide zero-valued or small activations~\citep{albericiocnvlutin2016,judd2017cnvlutin2}, and quantization during training~\citep{jain2018gist}. In any case, low-end devices with fewer compute resources may not even benefit from weight reuse across batches, as training on small batches and single images is often effective~\citep{masters_revisiting_2018}.

In this paper, therefore, we focus on reducing the number of \emph{weights} that must be stored during the training process. This is possible because deep networks have many more parameters than are needed to capture the intrinsic complexity of the tasks they are being asked to solve~\cite{li2018measuring}. Importantly, training only a fraction of the weights to change naturally results in a pruned model \emph{without} the need to refine the pruned weight set via additional training.

Our algorithm, \ours{}, is a training-time pruning technique that greedily selects the most promising subset of weights to train. It relies on three observations:
\setenumerate{label=(\alph*)}
\begin{enumerate}
\item that the parameters that have accumulated the highest total gradients account for most of the learning,
\item that the values of these parameters are predicated on the initialization values chosen for the remaining parameters, and
\item that initialization values can be cheaply recomputed on-the-fly without needing to access memory.
\end{enumerate}

In contrast to prior pruning techniques, which train an unconstrained network, prune it, and retrain the pruned network, \ours{} prunes the network immediately at the start of training and does not require a retraining pass. Only the weights that remain require memory, and most of the weight set is never stored. Because \ours{} recomputes initialization parameters, it can prune layers like batch normalization or parametric ReLU, which are not pruned in existing approaches.

\ours{} outperforms best-in-class pruning-only methods (which require retraining) on network architectures that are already dense and have been found particularly challenging to compress~\citep{liu_learning_2017,louizos_learning_2017,li_training_2017}. On Densenet, we achieve 5.86\% validation error with 4.5$\times$ weight reduction (vs.\ best prior 5.65\% / 2.9$\times$ reduction), and on WRN-28-10 we achieve 3.85\%--4.20\% error with 4.5$\times$--7.3$\times$ weight reduction (vs.\ best prior 3.75\% uncompressed and best pruned prior 16.6\% / 4$\times$ reduction). Using ResNet18 on ImageNet, we observe an 11.7$\times$ weight reduction without accuracy loss, and up to 24.4$\times$ with a small accuracy impact. Like other pruning techniques, it is orthogonal to, and can be combined with, quantization and value compression.

\section{\ours{}: training pruned networks}

\subsection{Approach and key insights}

Existing deep neural network pruning techniques rely on the observation that most tasks can be solved using many fewer parameters than the total size of the model~\cite{li2018measuring} — in other words, many parameters do not contribute useful information to the final output, and can be removed. Because the exact values of the surviving weights still depend on the values of the zeroed parameters, the network must still be retrained; only after retraining is it possible to recover close to original accuracy with an order of magnitude fewer weights~\citep[etc.]{handeep2015,zhusparsenn:2017, gecompressing2017, masanadomain-adaptive2017, luothinet:2017, ullrichsoft2017, EnergyAwarePruning:CVPR:2017}. While they reduce the energy expended on memory accesses during inference, they require retraining and therefore actually \emph{increase} the number of energy-consuming memory accesses during training.

\ours{} has a different purpose: we aim to reduce the memory footprints both \emph{during} and \emph{after} training, and so we must reduce the \emph{number of parameters tracked at training time}. To choose which parameters to track, we are guided by three related but slightly different insights, described below.

\begin{figure}
\centering
\includegraphics[width=\linewidth]{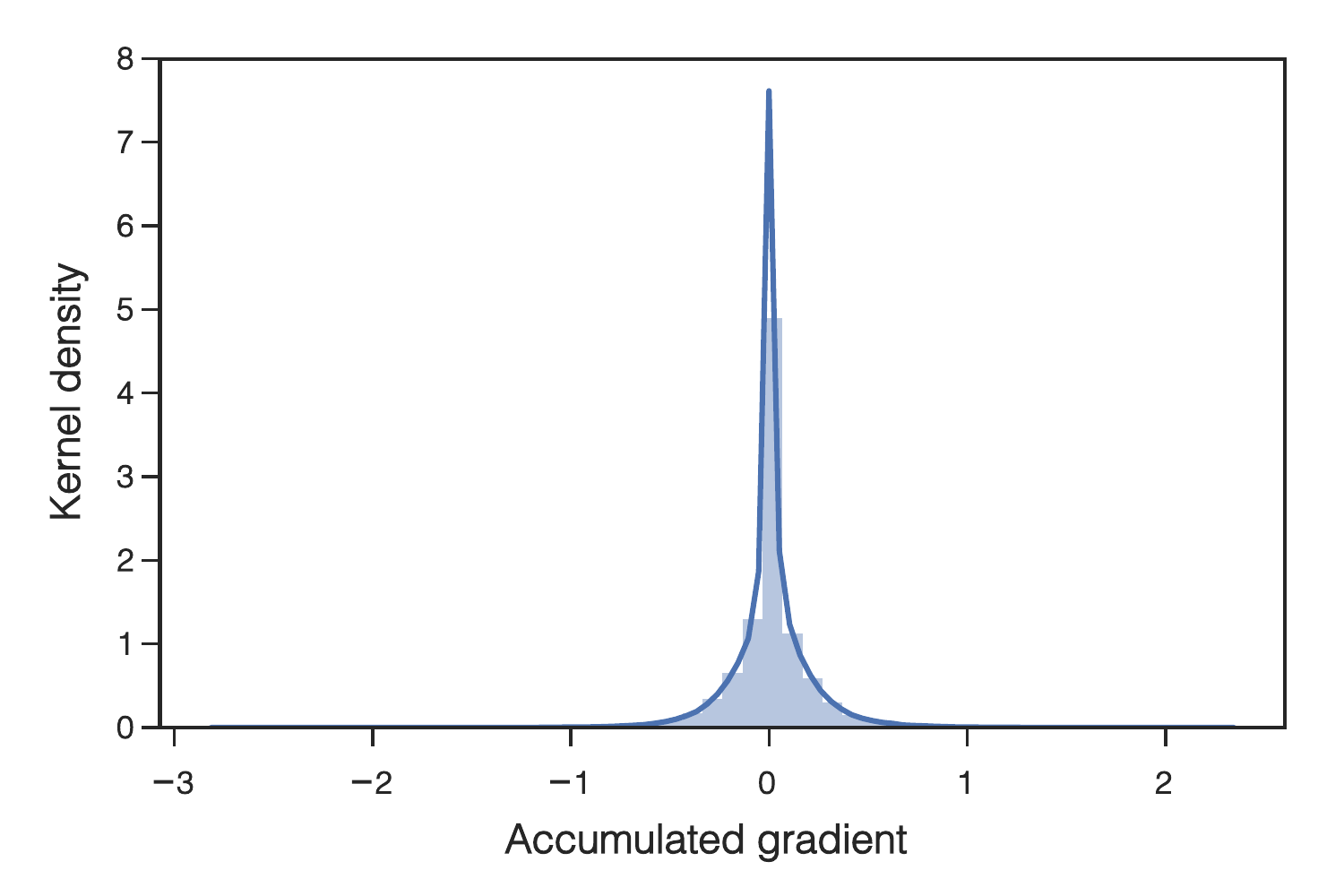}
\vspace{-6ex}
\caption{Distribution of accumulated gradients over 100 epochs of standard SGD training on MNIST using a 90,000-weight MLP.}
\label{fig:graddist}
\end{figure}

\begin{figure*}
\centering
\includegraphics[width=0.495\linewidth]{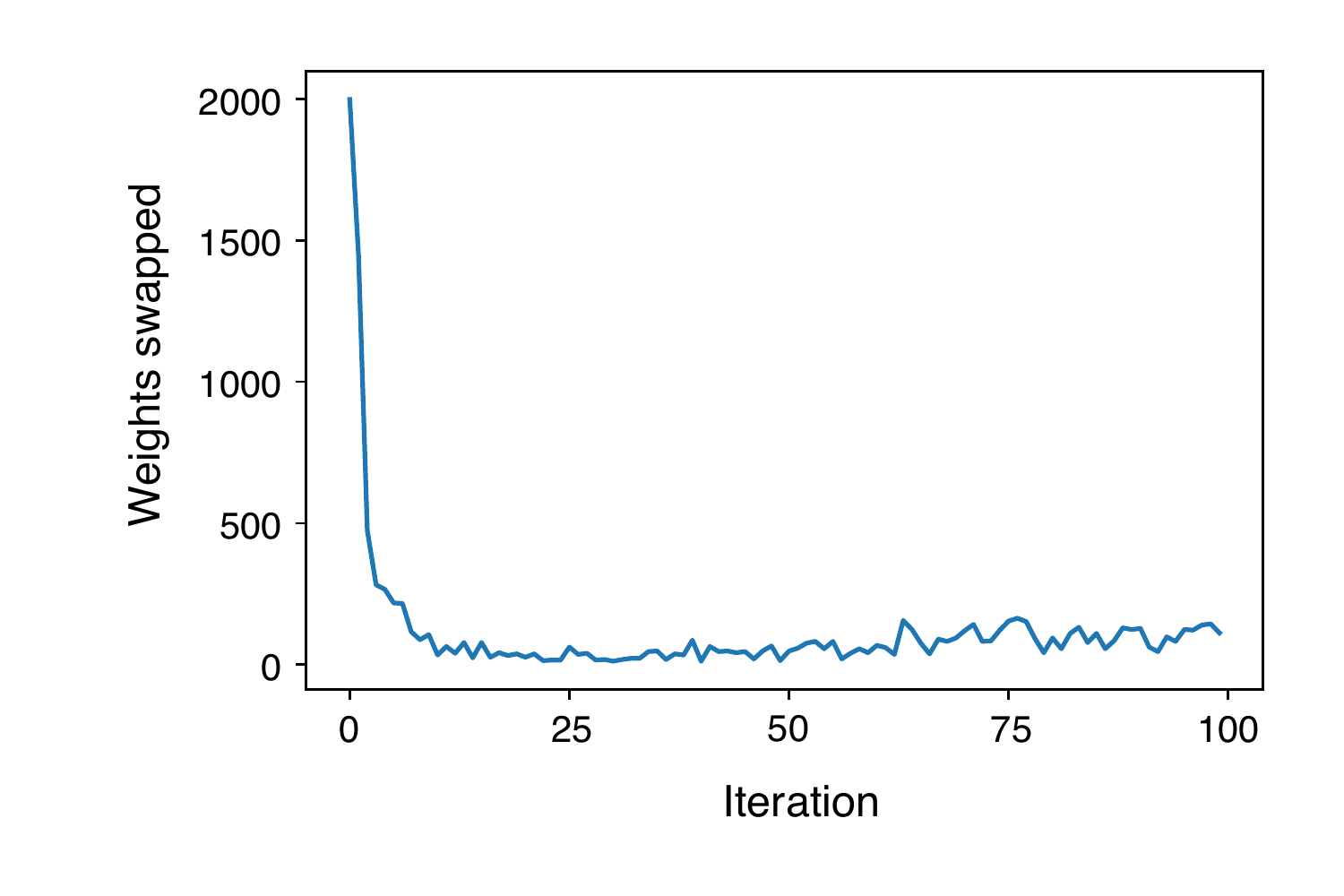}
\includegraphics[width=0.495\linewidth]{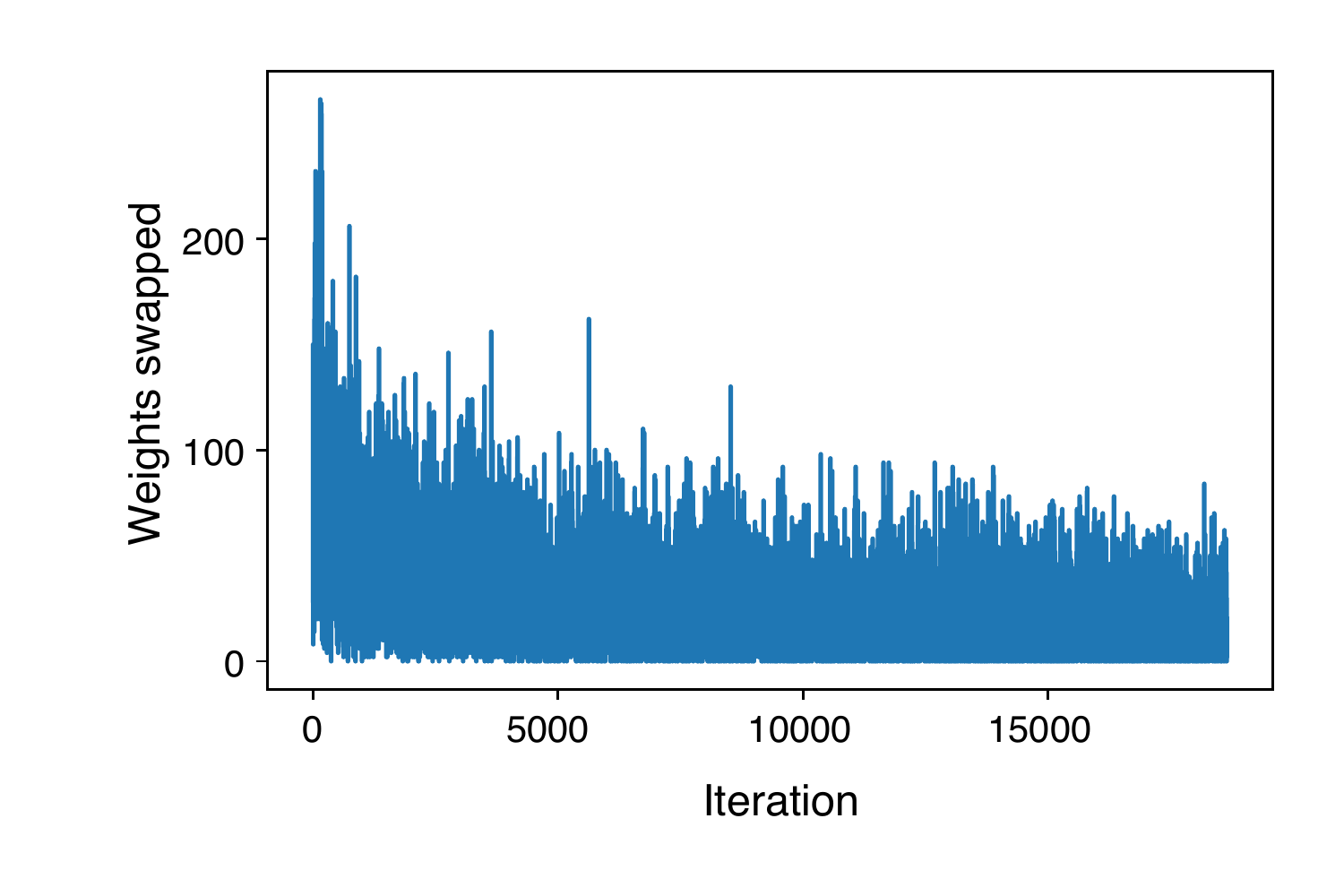}
\vspace{-4ex}
\caption{The number of weights added/removed to the top-2K gradient set in the first 10 mini-batches (left) and the remaining mini-batches (right) on MNIST. (Note different y-axis scales). }
\label{fig:drops}
\end{figure*}

\paragraph{Track the highest accumulated gradients.}  As post-training pruning techniques delete weights with the lowest \emph{values}, it may seem natural to keep track of gradients for the weights with the highest values. However, this naïve approach is not effective during the first few training iterations. Informally, this is because the initial value of each weight, typically drawn from a scaled normal distribution~\citep{lecuneffbackprop1998}, serves as scaffolding which gradient descent can amplify to train the network. To overcome this, pruning algorithms typically retrain the network after the lowest-value weights have been removed~\citep[etc.]{handeep2015,zhusparsenn:2017, gecompressing2017, masanadomain-adaptive2017, luothinet:2017, ullrichsoft2017, EnergyAwarePruning:CVPR:2017}.

Instead, our approach is to track gradients for a fixed number of weights which have learned the most overall — that is, the weights with the \emph{highest accumulated gradients}. \autoref{fig:graddist} shows that most weights move very little from their initial values and most accumulated gradients are near 0, suggesting that we only need to track gradients for a small fraction of the weights — \emph{provided} we keep the remaining weights at their initial values.

Note that the set of weights with the highest accumulated gradient may (and does) change as the training process explores different subspaces. \ours{} allows weights to enter the tracked set if their gradient exceeds the accumulated gradient of a currently-tracked weight; in that case, the weight with the lowest accumulated gradient is evicted from the tracked set and its value is reset to its initialization value.

With untracked weights kept at their initialization values, this approach results in a weight diffusion that is very close to that of the baseline SGD, a key factor in maintaining good generalization under different training regimes~\citep{diffusionpaper}.

\paragraph{Recompute initialization-time values for untracked weights.} While it may be intuitive to simply set untracked weights to zero, we observed that preserving the scaffolding provided by the initialization values is critical to the accuracy of the trained network. In our experiments on MNIST, we were able to reduce the tracked weights 60$\times$ if initialization values were preserved, but only 2$\times$ if untracked weights were zeroed. This is in line with the fact that prior pruning approaches require retraining after a subset of weights have been set to zero~\citep[etc.]{handeep2015,zhusparsenn:2017, gecompressing2017, masanadomain-adaptive2017, luothinet:2017, ullrichsoft2017, EnergyAwarePruning:CVPR:2017}.

However, storing the initialization-time values would require accessing off-chip memory to retrieve all of them during both the forward and backward passes of training — a costly proposition when a memory access consumes upwards of 700$\times$ more energy than a floating-point operation.

To avoid storing these weights, we observe that in practice the initial values are generated using a pseudo-random number source that is initialized using a single seed value and postprocessed to fit a scaled normal distribution. Because each value only depends on the seed value and its index, it can be deterministically regenerated exactly when it is needed for computation, without ever being stored in memory.\footnote{In a CPU or GPU, short-lived temporary values are stored in on-chip register files or scratchpads, reading which costs a fraction of the energy needed to access off-chip DRAM.} For example, recomputing a normally distributed pseudo-random initialization value using the xorshift algorithm~\citep{xorshift:2003} requires six 32-bit integer operations and one 32-bit floating point operation; this amounts to about 1.5pJ in a 45nm process, 427$\times$ less energy than a single off-chip memory access.

Regenerating untracked parameters to their initial values also works out-of-the-box for layers like Batch Normalization or Parametric ReLU, where the initialization strategy is typically a constant value (xorshift is not used for these). These layers are also pruned by \ours{}, which to the best of our knowledge is unique.

\paragraph{Freeze the set of tracked weights after a few epochs.} During training, gradients are still \emph{computed} for the untracked weights, and those gradients can exceed the accumulated gradients of any weights that are being tracked; this is especially likely during the initial phases of training, when the optimization algorithm seeks the most productive direction. While the energy needed to compute the gradient is not significant, replacing the lowest tracked gradients with newly computed ones would require additional memory references, which in turn would expend additional energy.

Once the network has been trained for a few epochs, however, we would expect the accumulated gradients for the tracked weights to exceed any “new” gradients that could arise from the untracked weights. To verify this intuition, we trained a 90,000-parameter MLP on MNIST using standard SGD while keeping track of which parameters were in the top-2K gradients set. \autoref{fig:drops} shows that the set of the highest-gradient weights stabilizes after the first few iterations. The “noise” of less than $0.04\%$ of weights entering and leaving the highest-gradient set in the rest of the epochs remains throughout the training process, and has no effect on the final accuracy. This observation allows us to freeze the “tracked” parameter set after a small number of epochs, saving more energy-costly memory accesses.

\subsection{The \ours{} algorithm}

\autoref{alg:by-accgrad-drng} shows the resulting \ours{} training process. 

\RestyleAlgo{algoruled}
\begin{algorithm}
\DontPrintSemicolon
\textbf{Initialization:} $W^{(0)} \text{~with~} W^{(0)} \sim{} N\!\left(0, \sigma\right)$\;
\textbf{Output:} $W^{(t)}$\;
	\While{not converged}  {
		$T = \left\lbrace\Big|\sum_{i=0}^{t-1} \frac{\alpha \partial{}f\left(W^{(i-1)}; x^{(i-1)}\right)}{\partial w} \Big| \text{~s.t.~} w \in W_\mathit{trk} \right\rbrace$\;
		if not frozen: \;
		\mbox{\hspace{5.4ex}} $U = \left\lbrace\Big|\frac{\alpha \partial{}f\left(W^{(i-1)}; x^{(i-1)}\right)}{\partial w} \Big| \text{~s.t.~} w \in W_\mathit{utrk} \right\rbrace$\;
		else: \;
		\mbox{\hspace{5.4ex}}$U =  \left\lbrace\right\rbrace$\;
		$S = \operatorname{sort}\!\left(T \cup U\right)$\;
		$\lambda = S_{k}$\;
		$\mathit{mask} = \mathbbm{1}\!\left(S > \lambda\right)$\;
		$W^{(t)} = \mathit{mask} \,\cdot{} \left(W^{(t-1)} - \alpha \nabla\!f\!\left(W^{(t-1)}; x^{(t-1)}\right)\right) + \overline{\mathit{mask}} \cdot{} W^{(0)}$\;
		$t = t+1$\;
	}
	 \caption{\ours{} training. $N\!\left(0, \sigma\right)$ is generated from the xorshift pseudo-RNG. $W_\mathit{trk}$ and $W_\mathit{utrk}$ = tracked and untracked weights; $T$ and $U$ = tracked and untracked accumulated gradients; $S$ = sorted accumulated gradients; $k$ = number of gradients to track and $\lambda$ = lowest tracked cumulative gradient; $\alpha$ = learning rate. $\mathit{mask}$ indicates a boolean matrix with the same shape as the weights. $\overline{\mathit{mask}}$ indicates the logical inverse of the mask.}
\label{alg:by-accgrad-drng}
\end{algorithm}

Weights are initialized from a scaled normal distribution~\mbox{\citep{lecuneffbackprop1998}}, computed via the xorshift pseudo-random number generator~\citep{xorshift:2003}. In every iteration, gradients are computed and the highest $k$ accumulated gradients are stored and carried to the next iteration.

For clarity of exposition, \autoref{alg:by-accgrad-drng} shows all gradients as recomputed and sorted to determine the highest-total-gradient set of $k$ elements. In a practical implementation, however, the tracked accumulated gradient set is stored a priority queue of size $k$ where the minimum elements are evicted when incoming gradients are higher than the stored minimum. Note that it suffices to store only the tracked set $T$: $W^{t}$ can be computed from $T$ and $W^{(0)}$, and $W^{(0)}$ can be determinstically regenerated when needed using the weight index and the xorshift random-number generator.

After a manually chosen iteration cutoff, the tracked set is “frozen” and gradients are only computed and updated for the weights already tracked; this saves additional computation time and energy. As in standard stochastic gradient descent, training ends once the network is considered to have converged.

\paragraph{\ours{} versus sparsity-based regularization.}
\ours{} can be contrasted with techniques like Dropout~\citep{srivastavadropout:2014} or DSD~\citep{handsd:2017}, which \emph{temporarily} restrict the gradients that can be updated, essentially as a regularization scheme. Dropout restricts randomly selected gradient updates during each training iteration, while DSD repeatedly alternates sparse phases (where the lowest-absolute-value weights are deleted) and dense refinement phases (where all weights may be updated). In both cases, the training process is allowed to update all weights much of the time, even though a only a subset of weights may update in a specific phase of training.

In contrast, \ours{} limits the set of weights that can be updated \emph{throughout} the entire training process — there is never a phase where non-tracked weights are updated. Weights that have not substantially contributed to the overall optimization gradient are not trained, and retain their initialization values. Because the final performance of a \ours{}-trained network depends only on the initialization values and the accumulated gradients of the tracked subset of weights, retraining is not needed.

\section{Experiments}
\label{sec:results}

\subsection{Methods} We implemented \ours{} using the Chainer deep neural network toolkit~\citep{tokui2015chainer}; models were trained on an NVIDIA 1080Ti GPU. We compared \ours{} to a baseline implementation without any pruning, as well as three representative pruning techniques:
\begin{enumerate}
\item a straightforward magnitude-based pruning implementation where only the highest weights are kept after each iteration,
\item variational dropout~\citep{kingma_variational_2015}, which can progressively remove weights during training, and
\item network slimming~\citep{liu_learning_2017}, a modern train-prune-retrain pruning method that achieves state-of-the-art results on modern network architectures.
\end{enumerate}
All networks were optimized using stochastic gradient descent with momentum, as all other optimization strategies cost significant extra memory. 

		\begin{table*}[t]
		\centering
		\begin{tabular}{llllll}
		\toprule
		MNIST-300-100 & Validation Error & Weight reduction & Best Epoch & Freeze Epoch   \\
		\midrule
		Baseline 267K & 1.41\%           & 0$\times$     & 65 & N/A  \\
		\ours{} 50K & 1.51\%           & 5.33$\times$       & 24    & 100 \\
		\ours{} 5K  & 2.58\%           &  53.32$\times$      & 32  & 20 \\
		\ours{} 1.5K  & 3.84\%           &  177.74$\times$      & 97  & 40\\
		\midrule
		MNIST-100-100& Validation Error & Weight reduction & Best Epoch & Freeze Epoch   \\
		\midrule
		Baseline 90K& 1.70\%           & 0$\times$          & 47 & N/A  \\
		\ours{} 50K & 1.58\%           & 1.8$\times$       & 24    & 5 \\ 
		\ours{} 20K & 1.70\%           & 4.5$\times$       &  32  &  5\\ 
		\ours{} 1.5K  & 3.78\%           & 60$\times$      & 26  & 30 \\ 
		\bottomrule
		\end{tabular}
		\caption{The MNIST digit dataset using LeNet-300-100 (top) and MNIST-100-100, a smaller MLP with 100 hidden neurons (bottom). \ours{} 50K refers to a configuration where 50,000 gradients are retained during training, \ours{} 5K refers to a configuration with 5,000 retained gradients, and so on.}
		\vspace{-2ex}
		\label{table:mnist}
\end{table*}

\begin{figure*}
\centering
\includegraphics[width=0.495\linewidth]{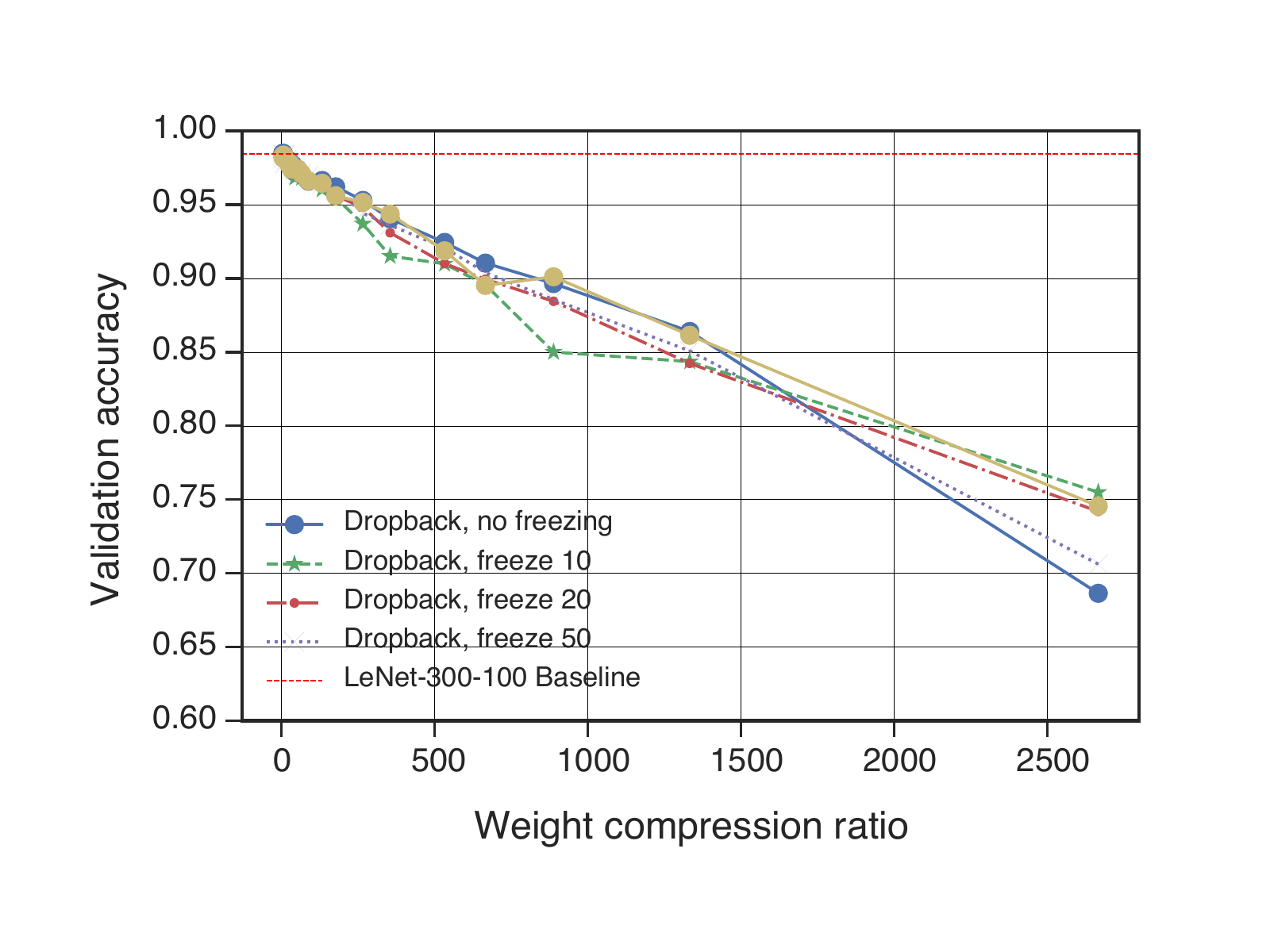}
\includegraphics[width=0.495\linewidth]{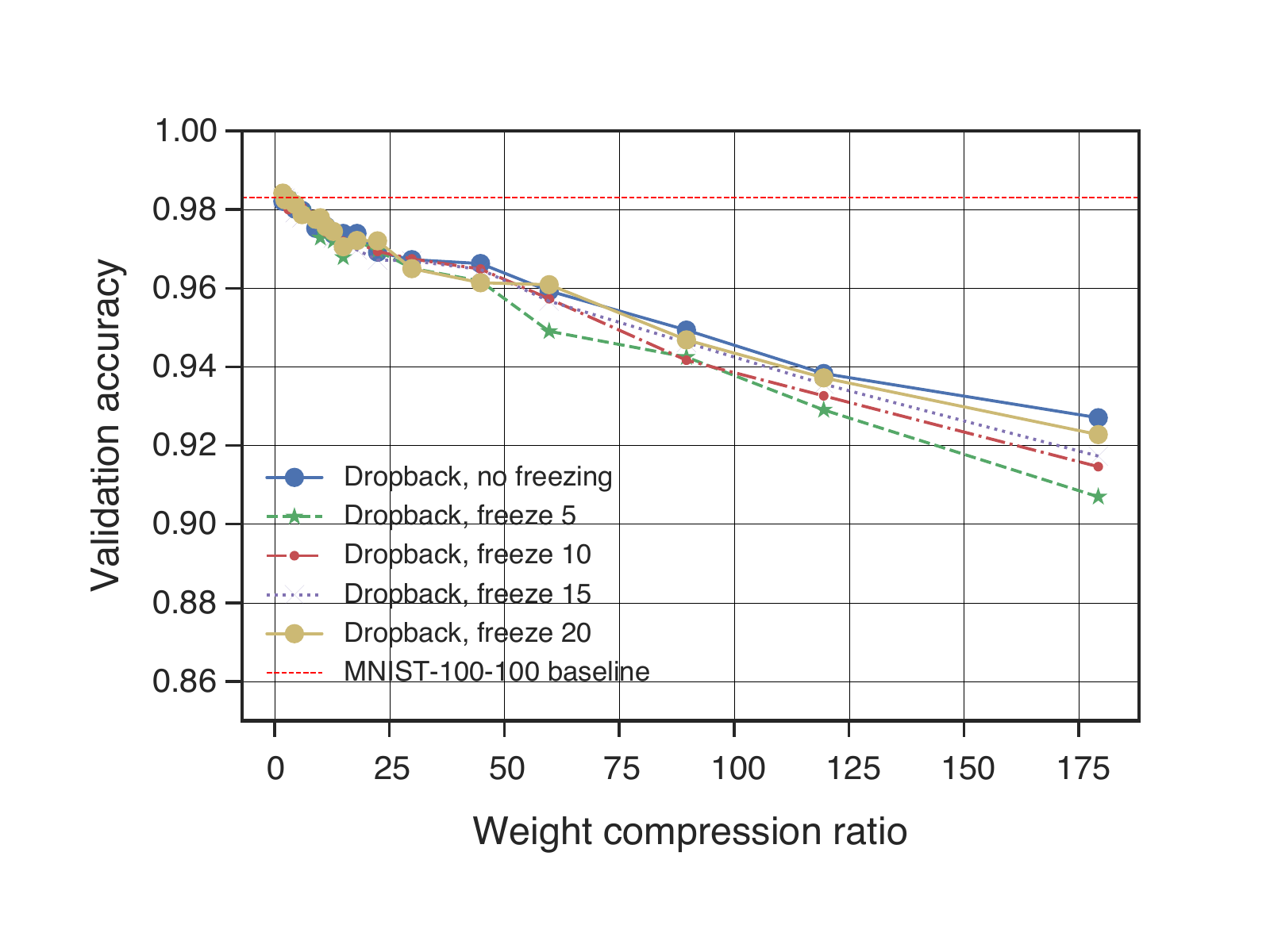}
\vspace{-6ex}
\caption{Tracked parameter tradeoff vs. network accuracy, with several parameter selection freezing points for LeNet-300-100 (left) and MLP-100-100 (right) on MNIST. Freeze 0 represents a network trained without freezing.}
\label{fig:lemnist}
\end{figure*}

We evaluated all techniques on the MNIST~\citep{lecun_mnist_1998}, CIFAR-10~\citep{Krizhevsky2009LearningML}, and ImageNet~\citep{Russakovsky:ImageNet:2015} datasets. For MNIST, we used both LeNet-300-100~\citep{lecun_gradient-based_1998} and a simpler network with only 100 hidden units, which we refer to as MNIST-100-100. For CIFAR-10, we used three networks: Densenet~\citep{huang_densely_2016}, WRN-28-10~\citep{zagoruyko_wide_2016}, and VGG-S, a reduced VGG-16-like model with dropout, batch normalization, and two FC layers of 512 neurons including the output layer (a total of 15M parameters vs.\ the 138M of VGG-16). We specifically chose Densenet and WRN because they represent modern network architectures that are very challenging to prune with existing techniques~\citep{liu_learning_2017,louizos_learning_2017,li_training_2017}. For ImageNet, we trained the ResNet18 model~\citep{hedeep2016}; as at the time of writing we have been unable to access the test set evaluation server for ImageNet, we report all results on the validation set.

In what follows, we report reductions in network size as the ratio of the unpruned weight count to the number of weights tracked during training. Note that this number refers to weight pruning only, and does \emph{not} include other reduction techniques such as quantization or compression, which are orthogonal to our approach.

\subsection{MNIST digit recognition} We first evaluated \ours{} on the MNIST handwritten digits dataset using a small multi-layer perceptron (MLP) with approximately 90,000 weights, as well as the LeNet-300-100 MLP, which has approximately 266,600 weights. Training was allowed for up to 100 epochs, and the initial learning rate of 0.4 was cut in half four times during training. The best epoch was chosen as the epoch with the highest validation accuracy after five epochs of no improvement.

\paragraph{Weight reduction and accuracy.} \autoref{table:mnist} shows the results for the baseline (unpruned) model and three configurations of \ours{}, retaining respectively 50,000 weights (1.8$\times$ reduction), 20,000 weights (4.5$\times$ reduction), and 1,500 weights (60$\times$ reduction). With MNIST-100-100 and a modest 2$\times$ reduction in weights, \ours{} slightly exceeds the accuracy of the baseline model. This matches the trend reported in train-prune-retrain work such as DSD~\citep{handsd:2017}, where a sparse model that omits 30\%–50\% weights outperforms the baseline dense (all-weights) model. The larger MLP sees a slight drop in accuracy, but at 50,000 tracked weights is compressed many more times, and reaches maximum accuracy nearly 3$\times$ faster. Further reducing the model to 20,000 weights (4.5$\times$ reduction) results in nearly the same accuracy as the baseline, and convergence in a comparable number of epochs. 

\begin{figure}
\centering
\includegraphics[width=\linewidth]{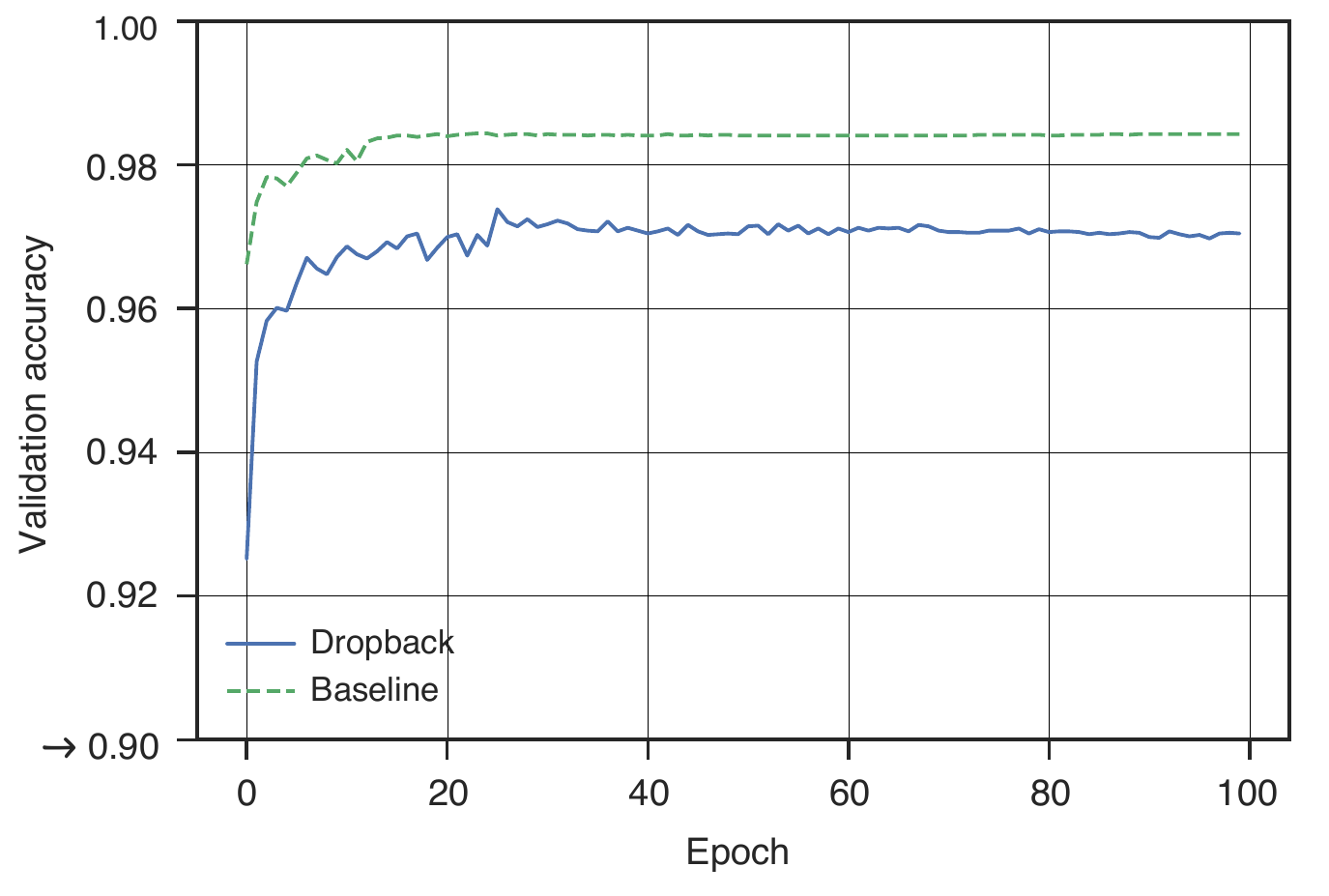}
\vspace{-4ex}
\caption{Rate of convergence for LeNet-300-100 for our technique our and the baseline model. Note that the y-axis starts as 0.90, and the accuracies are within 1\% of each other.}
\label{fig:lenetconv}
\end{figure}

		\begin{table*}[h!]
		\centering
		\begin{tabular}{lllll}
		\toprule
		layer          & \ours{} 1500 & \ours{} 10000 & Baseline \\
		\midrule
		fc1 (100$\times$784)     & 734 (52.3\%)    & 7223   (72.2\%)   & 78500  (87.6\%)   \\
		fc2 (100$\times$100)     & 512   (34.1\%)   & 2128    (21.3\%)  & 10100   (11.3\%)  \\
		fc3 (100$\times$10)      & 254   (16.9\%)   & 549    (5.5\%)   & 1010   (1.1\%)   \\
		\midrule
		Total & 1500  & 10000 & 89610\\
		\bottomrule
		\end{tabular}
		\caption{{}Number of gradients for each layer retained in the final trained MNIST network}
		\label{table:mnistWeights}
		\end{table*}

We also investigated an extreme reduction configuration with weight storage reduced drastically to 1,500 weights. Not surprisingly, the error rate increases (over 2$\times$) and twice the number of epochs are needed to achieve convergence. Still, the nearly 60$\times$ reduction in the weight count for the smaller model and 177$\times$ for the larger model potentially offer an attractive design point for low-power embedded accelerators in future mobile and edge devices.

\paragraph{Tracked weight set freezing.} \autoref{fig:lemnist} shows the effect of different freezing times for the tracked gradient set versus the tracked gradient counts. For both MNIST networks, freezing sooner to reduce the computational overhead results in lower achieved accuracy — especially for very high reduction ratios — but for smaller reduction ratios freezing early has little effect on the overall accuracy. \autoref{fig:lenetconv} shows the rate of convergence for \ours{} and the baseline on the LeNet-300-100 network; despite different tracked parameter counts, both methods have similar convergence behaviour.

\paragraph{Retained weight distribution.} \autoref{table:mnistWeights} shows that the number of parameters retained per layer varies depending on the number of tracked weights. The smaller \ours{} 1.5K network allocates a much higher amount of its weights to the later layers compared to the \ours{} 10K network and the baseline — an indication that, in the smaller network, proportionally more neurons in the later layers are critical for decision-making.

\subsection{CIFAR-10 image classification} We also studied the training performance of \ours{} using VGG-S, Densenet, and WRN-28-10 on the CIFAR-10 dataset. This is a much more challenging task than MNIST, and the networks were specifically selected for being already relatively dense. Models were trained for 300 epochs on VGG-S and 500 epochs on Densenet and WRN; the best epoch was selected. No data augmentation of CIFAR-10 was performed. The learning rate was initialized to 0.4 and decayed 0.5$\times$ every 25~epochs. 

\begin{table*}[]
\centering
		\begin{tabular}{llllll}

		\toprule
		CIFAR-10    & Validation error & Weight reduction & Best epoch  & Freeze epoch \\
		\midrule
		Baseline 15M & 10.08\%           & 0$\times$           & 214 &   N/A	\\
		VGG-S \ours{} 5M & 9.75\%           & 3$\times$         &   127  & 5 	\\ 
		VGG-S \ours{} 3M & 9.90\%           & 5$\times$         &  128  & 20 		\\ 
		VGG-S \ours{} 0.75M  & 13.49\%           & 20$\times$        & 269 & 35 	\\ 
		VGG-S \ours{} 0.5M  & 20.85\%           & 30$\times$        & 201 & 15 	\\ 
		VGG-S Var. Dropout & 13.50\%           & 3.4$\times$      & 200 & N/A 	\\  
		VGG-S Mag. Pruning .80 & 9.42\%           & 5.0$\times$      & 182 & N/A 		\\
		VGG-S Slimming & 11.08\%           & 3.8$\times$      & 196 & N/A 		\\
		\midrule
		Densenet Baseline 2.7M & 6.48\%           & 0$\times$      & 382 & N/A 		\\ 
		Densenet \ours{} 600K & 5.86\%           & 4.5$\times$      & 409 & N/A 		\\
		Densenet \ours{} 100K & 9.42\%           & 27$\times$      & 307 & N/A 			\\
		Densenet Var. Dropout & FAIL           & FAIL      & FAIL & FAIL 			\\
		Densenet Mag. Pruning .75 & 6.41\%           & 4.0$\times$      & 480 & N/A 		\\
		Densenet Slimming & 5.65\%           & 2.9$\times$      & N/A & N/A 		\\
		\midrule
		WRN-28-10 Baseline 36M & 3.75\%           & 0$\times$      & 326 & N/A 		\\ 
		WRN-28-10 \ours{} 8M & 3.85\%           & 4.5$\times$      & 384 & N/A 			\\
		WRN-28-10 \ours{} 7M & 4.02\%           & 5.2$\times$      & 417 & N/A 			\\
		WRN-28-10 \ours{} 5M & 4.20\%           & 7.3$\times$      & 304 & N/A 			\\
		WRN-28-10 Var. Dropout & FAIL           & FAIL      & FAIL & FAIL 			\\
		WRN-28-10 Mag. Pruning .75 & 26.52\%           & 4$\times$      & 109 & N/A 		\\
		WRN-28-10 Slimming .75 & 16.640\%           & 4$\times$      & 173 & N/A 			\\
		\bottomrule
		\end{tabular}
		\caption{{}Validation accuracy and reduction ratios on CIFAR-10. FAIL = training did not converge due to gradient explosion.}
		\label{table:cifar}

\end{table*}

\paragraph{Weight reduction and accuracy.} \autoref{table:cifar} shows how \ours{} compares to variational dropout, network slimming, and magnitude-based pruning on VGG-S, Densenet, and WRN-28-10. Overall, \ours{} is able to achieve comparable (or even slightly improved) accuracy on VGG-S and Densenet with five-fold weight reduction, and up to 20$\times$--30$\times$ if some accuracy is sacrificed. On WRN-28-10, \ours{} achieves 5$\times$ and 7$\times$ reduction with less than 0.5\% accuracy drop.

Note that these networks are challenging, as they are already quite dense for the accuracy level they achieve. Variational dropout works only on VGG-S, and fails to converge on Densenet and WRN due to gradient explosion. Magnitude-based pruning does not achieve better accuracy than \ours{} despite less weight reduction. Finally, network slimming achieves slightly better top accuracy on Densenet with 50\% less reduction, but results in dramatic accuracy loss when applied to WRN; this corresponds to recent work which has shown that WRN is hard to compress more than about 2$\times$ without losing significant accuracy~\citep{liu_learning_2017,louizos_learning_2017,li_training_2017}. \ours{}, in contrast, is universally able to achieve weight reduction on the order of 5$\times$ with little to no accuracy loss.

\begin{figure*}
\centering
\includegraphics[width=0.495\linewidth]{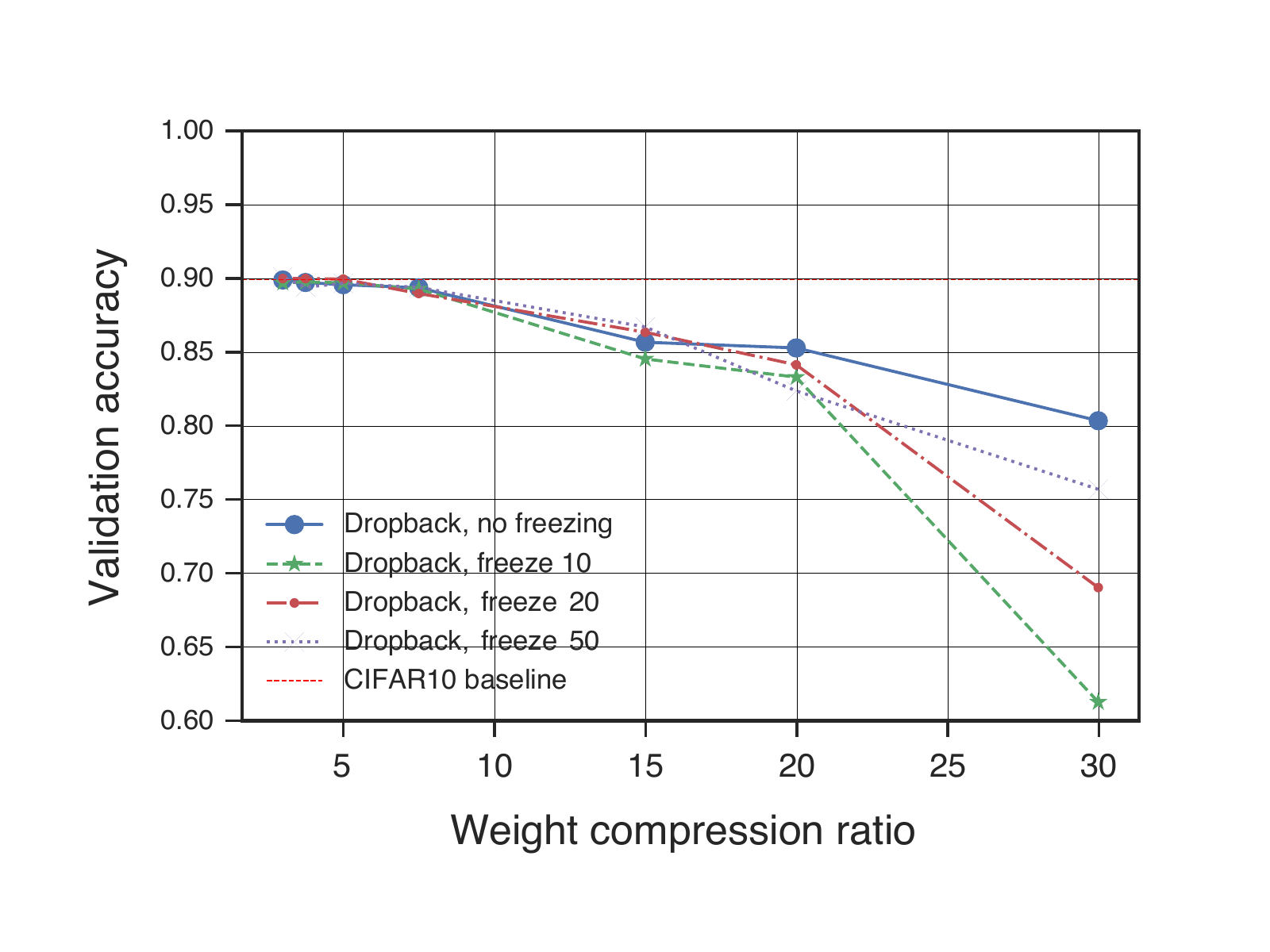}
\includegraphics[width=0.495\linewidth]{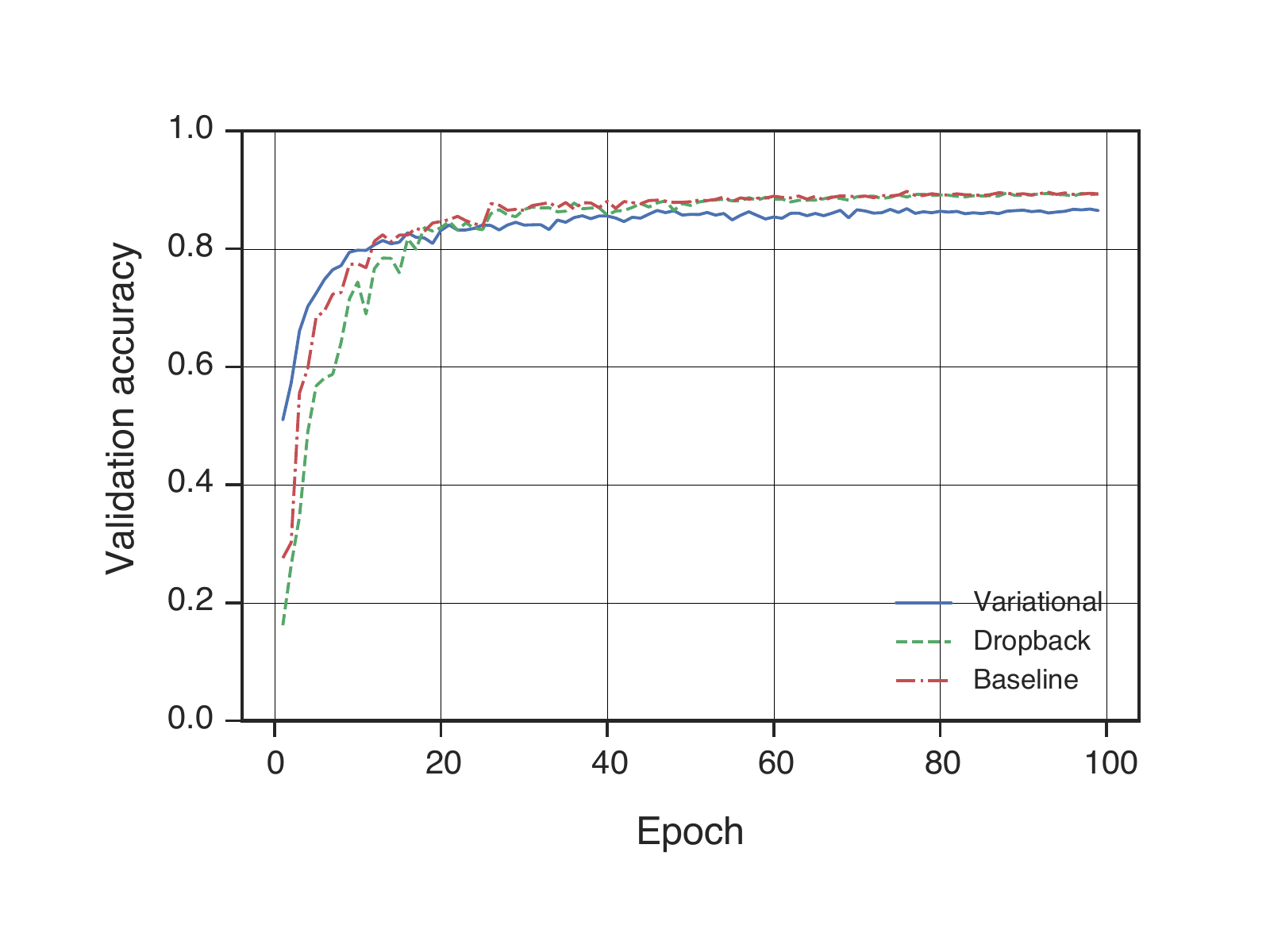}
\vspace{-5ex}
\caption{VGG-S CIFAR-10. Left: tracked parameter tradeoff vs. network accuracy, with several parameter selection freezing points (note the y-axis range). Right: epoch vs. validation accuracy for our method at 5M tracked parameters, variational dropout, and the baseline model.}
\label{fig:cifar}
\end{figure*}

\paragraph{Effects of freezing.} \autoref{fig:cifar}(right) shows how freezing the tracked parameter set after 10, 20, and 50 epochs affects the reduction and accuracy on VGG-S. At under $10\times$ weight pruning, accuracy does not suffer against the uncompressed baseline, and freezing does not affect either reduction or accuracy. With more reduction, however, the tracked gradient set struggles to maintain accuracy, and freezing the gradients early results in substantial accuracy drops at 30$\times$ reduction.

\paragraph{Convergence.} \autoref{fig:cifar}(left) shows that \ours{} initially learns slightly more slowly than the uncompressed baseline, but exhibits the same convergence behaviour after about 20 epochs (VGG-S). Variational dropout, in contrast, learns more quickly initially but converges on a substantially lower accuracy.

\subsection{ImageNet image classification}
Finally, we studied whether \ours{} can be applied in the more realistic setting of the ImageNet dataset. This dataset is substantially more challenging than CIFAR-10 — there are 1,000 instead of 10 categories and the images are 64$\times$ larger — and represents the high end of tasks for which inference is possible in a mobile-device setting.

ResNet18 was trained for 100 epochs using stochastic gradient descent with momentum, and the best epoch was selected.  The initial learning rate of 0.2 was decayed by 0.97$\times$ in each epoch. Freezing was not applied during \ours{} training, and no data augmentation was performed.

\begin{table*}[]
\centering
		\begin{tabular}{lllll}
		\toprule
		ResNet18 & Validation error & Weight reduction & Best epoch  \\
		\midrule
Baseline 11M & 31.59\%           & 1.00$\times$           & 26 	\\
DropBack 4M & 30.31\%           & 2.92$\times$           & 41 	\\
DropBack 2M & 29.95\%           & 5.85$\times$           & 44 	\\
DropBack 1M & 32.01\%           & 11.69$\times$           & 49 	\\
DropBack 0.5M & 39.53\%           & 23.39$\times$           & 47 	\\
Var. Dropout & FAIL & FAIL & FAIL \\
Mag. Pruning 2.75M & 38.43\%           & 4$\times$           & 6 	\\
		\bottomrule
		\end{tabular}
		\caption{Validation accuracy and reduction ratio of \ours{} applied to ResNet18 on the ImageNet dataset. \ours{} is able to reduce the weight count by 11.7$\times$ without accuracy loss, and nearly 24$\times$ with some degradation. In comparison, magnitude-based pruning at 4$\times$ reduction suffers the same loss as \ours{} at 23.4$\times$. All of our Variational Dropout experiments on ResNet18 failed due to numerical instability.}
		\label{table:resnet}

\end{table*}

\begin{figure*}
\centering
\includegraphics[width=0.495\linewidth]{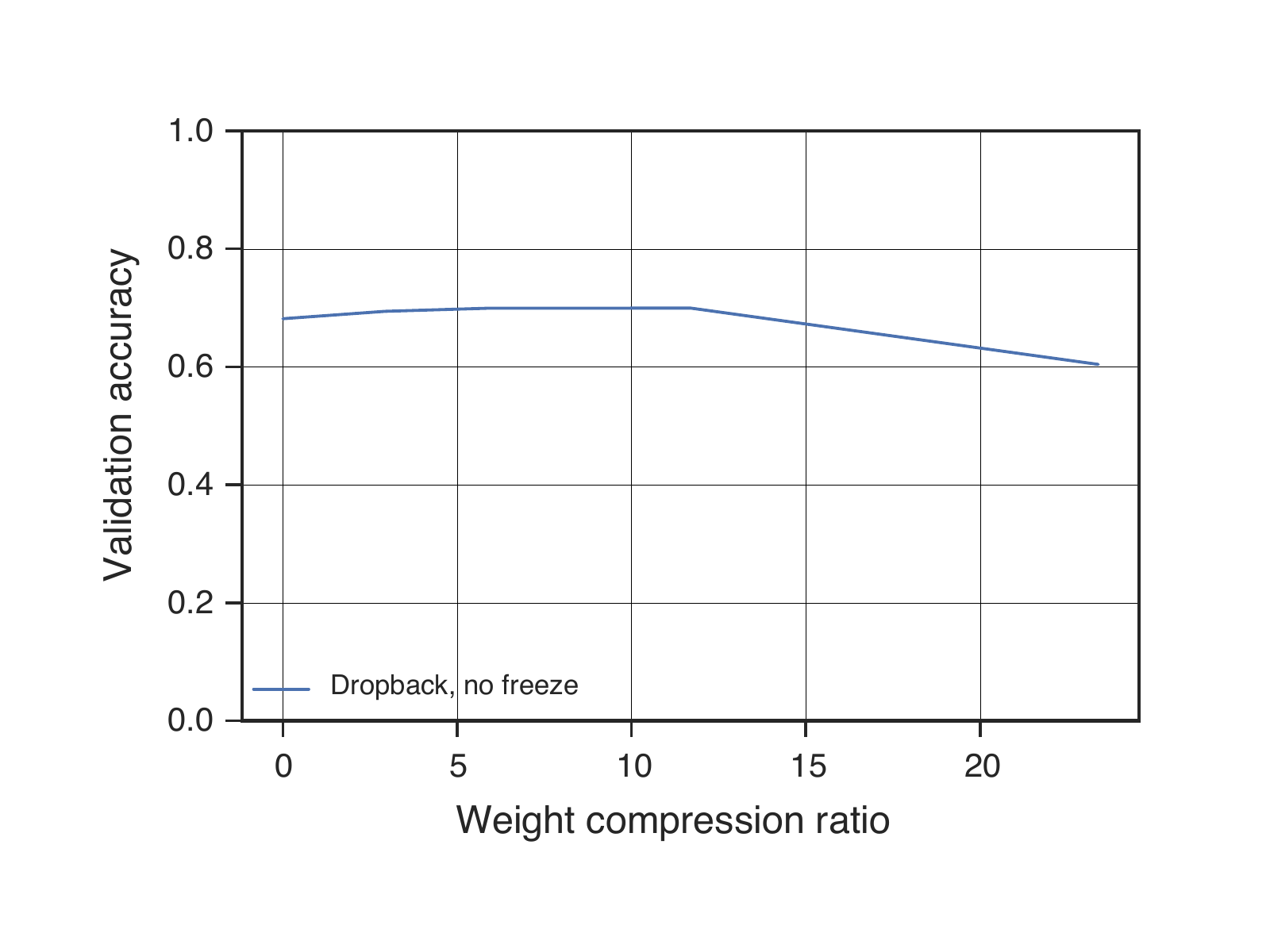}
\includegraphics[width=0.495\linewidth]{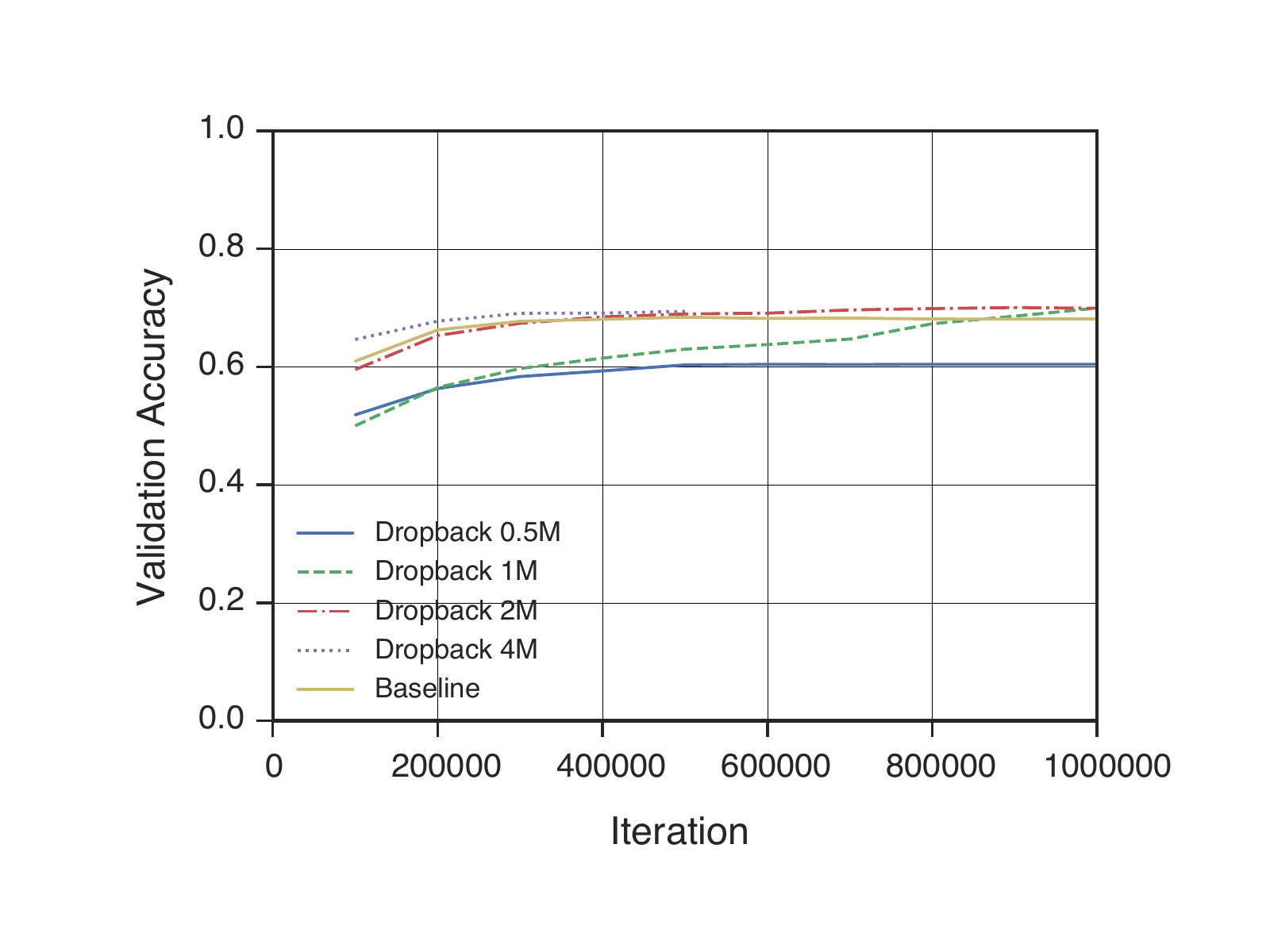}
\vspace{-4ex}
\caption{ResNet18 on ImageNet. Left: tracked parameter tradeoff vs.\ validation accuracy of the trained network. Right: epoch vs. validation accuracy at different weight reduction ratios, as well as the uncompressed baseline (which has 11M weights).}
\label{fig:res18}
\end{figure*}

\paragraph{Weight reduction and accuracy.} \autoref{table:resnet} and \autoref{fig:res18}(left) show top-1 results on the ImageNet dataset using the ResNet18 network (11 million weights uncompressed) for different weight reduction ratios. \ours{} is able to match or improve the accuracy of the baseline uncompressed network with weight reduction ratios of up to 11.7$\times$, and can reduce the weight count over 23$\times$ with a modest drop in accuracy (39.5\% vs 31.6\% baseline). In comparison, magnitude-based pruning stops improving after epoch 6 and never matches baseline accuracy: even a modest 4$\times$ reduction in weight count incurs the same cost in accuracy as \ours{} pays for a 23.4$\times$ reduction.

\paragraph{Convergence.} \autoref{fig:res18}(right) shows that \ours{} learns at the same rate as the uncompressed baseline up to 5.85$\times$ weight reduction. At 11.7$\times$ weight reduction, the model initially learns more slowly and takes substantially longer to converge, but eventually matches the baseline accuracy.

\section{Discussion}

\begin{figure}[t]
\centering
\includegraphics[width=\linewidth]{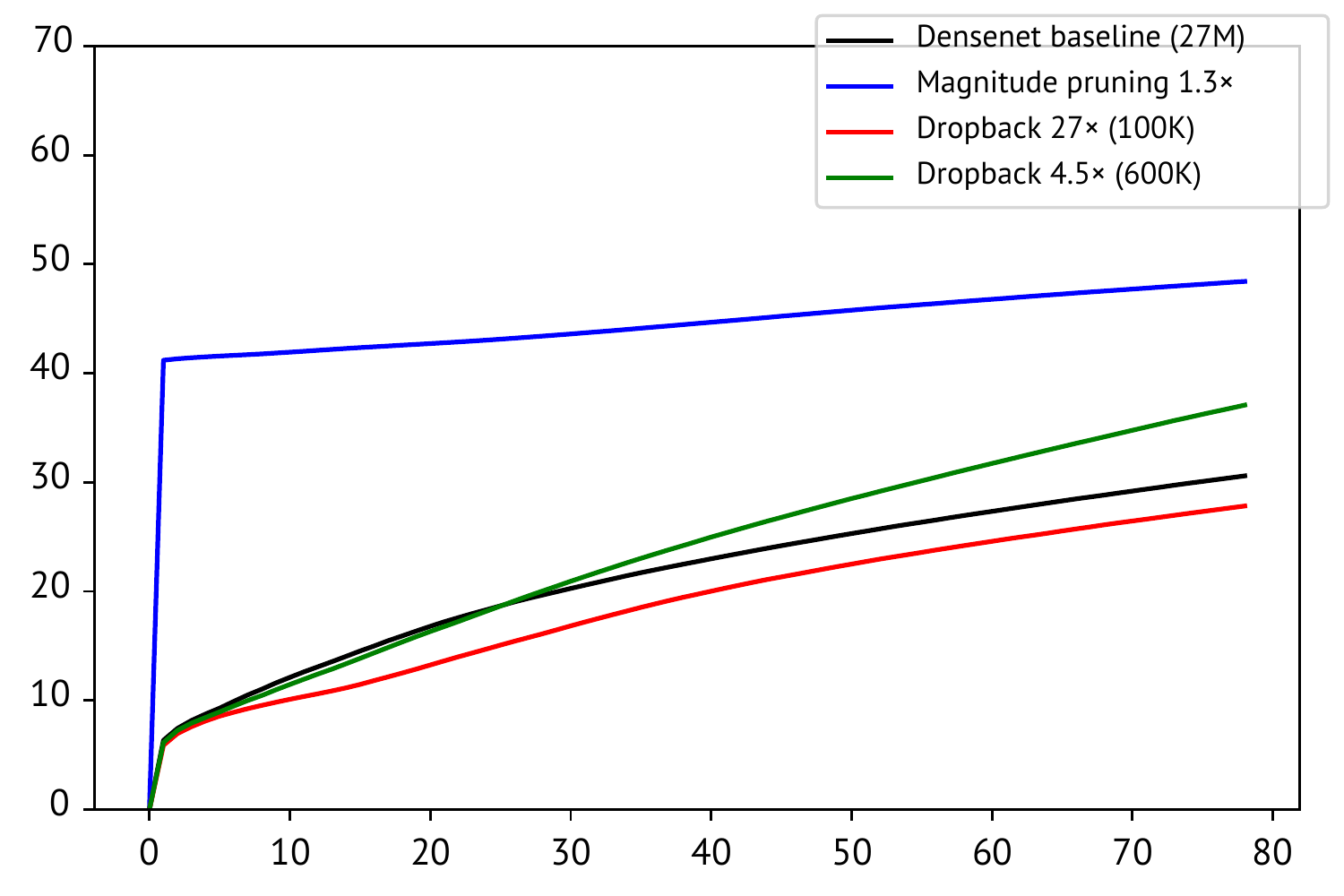}
\vspace{-4ex}
\caption{Diffusion ($\ell^2$) distance vs.\ training time for Densenet on CIFAR-10. Unlike magnitude-based pruning, \ours{} weight evolution follows an $\ell^2$ distance profile very similar to that of the uncompressed baseline throughout the training process.}
\label{fig:l2time}
\end{figure}

To investigate why \ours{} consistently achieves good reduction/accuracy tradeoffs across a wide range of networks, we considered the training process analysis due to~\citet{diffusionpaper}. Briefly, they observe that when DNNs are trained using SGD, the $\ell^2$ distance of weights $\mathbf{w}_t$ at time $t$ from the initial weights $\mathbf{w}_0$ grows logarithmically, i.e., \mbox{$\| \mathbf{w}_t - \mathbf{w}_0 \| \sim \log t$} (this is the same as the gradient accumulated until time $t$). They therefore model SGD as a random walk on a random potential surface, which exhibits the same logarithmic distance effect (known as ultra-slow diffusion). \citet{diffusionpaper} argue that SGD configurations (for them, batch sizes) that preserve the ultra-slow diffusion effect result in models that generalize well.

We reasoned that \ours{} maximally preserves the $\ell^2$ diffusion distance of the baseline training scheme because (a)~\ours{} tracks the highest gradients, and (b)~most of the remaining gradients are very close to zero (cf.\ \autoref{fig:graddist}). To verify this intuition, we measured the diffusion distance for the baseline uncompressed network, \ours{}, and magnitude-based pruning, all on the Densenet network classifying CIFAR-10.$\!$\footnote{Network slimming, being a train-prune-retrain technique, is not amenable to this type of analysis, and variational dropout failed to converge on that task.} \autoref{fig:l2time}(left) shows that under \ours{} weights diffuse similarly to the baseline training scheme, with the overall $\ell^2$ distance negligibly lower because the untracked weights remain at their initialization values.

To visualize how the weight values evolve under \ours{} compared to the baseline and pruning, we projected the parameter space down to 3D using principal component analysis (PCA). We repeated the experiment for \ours{}, baseline, and magnitude-based pruning; for each run, we saved weight values every 100 batches, and fit the weight history to a 3D PCA space. While each run starts with exactly the same initial weights, the PCA process maps these to different points in the 3D space depending on the full weight history; to clearly show the divergence of the training process, we translated all of the lines to align at the same initialization point in the post-PCA 3D space as well. 

\begin{figure}[t]
\centering
\includegraphics[width=\linewidth]{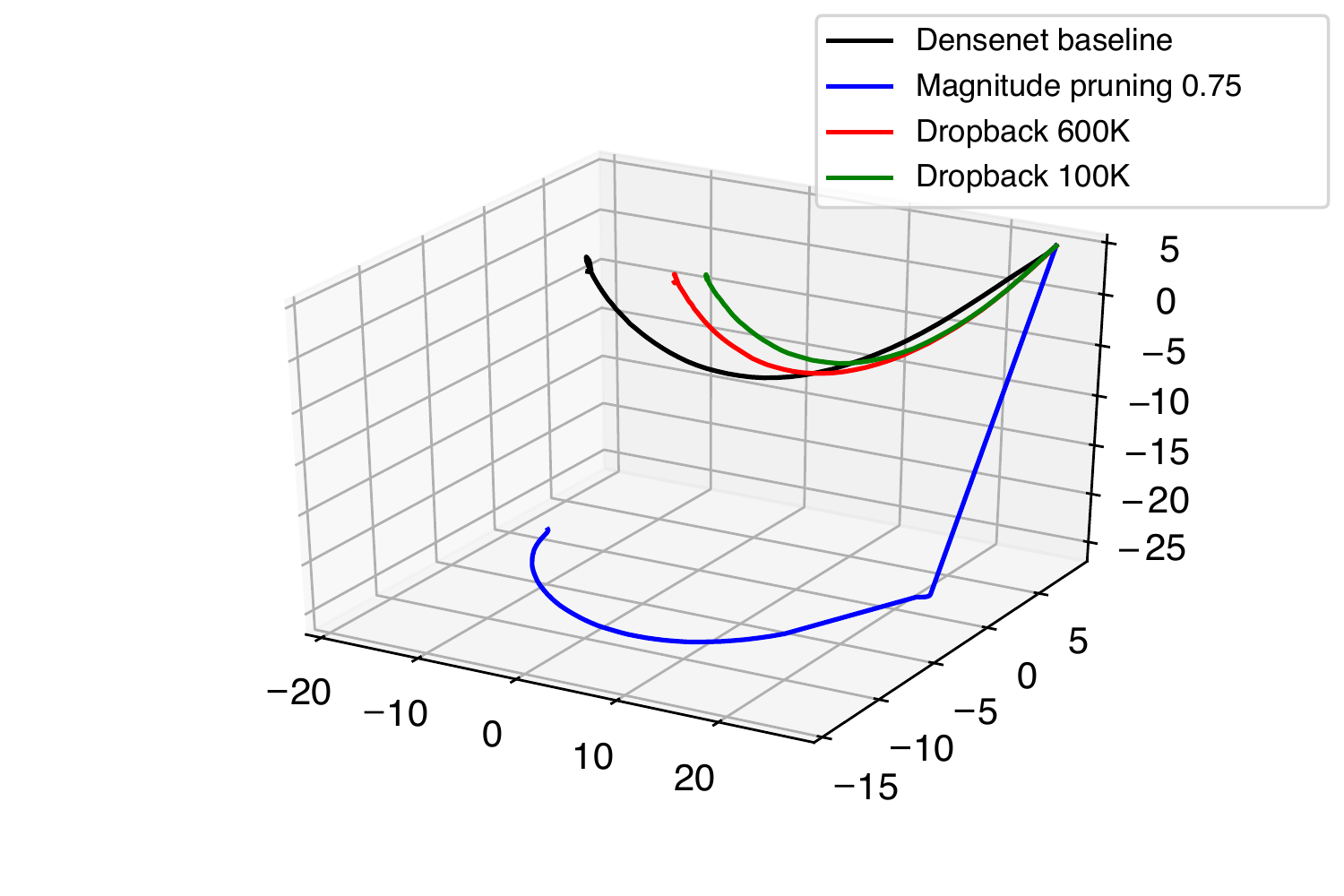}
\vspace{-4ex}
\caption{Evolution of weights under SGD projected into 3D space using PCA, training Densenet on CIFAR-10. To clearly show the training process, all projections were translated to align at their initial points. Unlike magnitude-based pruning, \ours{} training remains close to the uncompressed baseline.}
\label{fig:3d}
\end{figure}

\autoref{fig:3d} shows that under \ours{}, the principal components of the trained weight vector stay very close to those of the baseline-trained weight vector, whereas that of magnitude-based pruning diverges significantly. If we imagine the training path of the baseline uncompressed configuration to be optimal, \ours{} results in a close-to-optimal evolution.

\section{Related Work}

\subsection{Parameter and optimization landscape studies}
The idea that deep neural networks have more parameters than is necessary to solve the task at hand has been well established~\citep{dauphin2013big,denil2013predicting}. This observation has driven the development of pruning techniques (see below), and \ours{} relies on this insight as well.

\ours{} is also inspired by studies on optimization landscapes. \citet{Dauphin:2014:SaddlePoint} demonstrated that local critical points tend to be saddle-points, suggesting that optimization gradients are productive along many dimensions even at the critical point. Optimization paths that lead from the initial point to the optimum are often direct~\citep{goodfellow2014qualitatively}; this suggests that an algorithm like \ours{} that explores only a subspace of the optimization domain has a good chance of arriving at the optimum.

Closest in spirit to \ours{} is the study by~\citet{li2018measuring}, which trained a limited-dimensionality subspace of a full model and used random matrices to project the training subspace to the full model in an approach reminiscent of compressive sensing. Because the dimensions of the training subspace are \emph{not} a subset of the dimensions of the full parameter space, however, the subspace must still be projected to the full model for evaluation. Unfortunately, this approach is not suitable for training with limited resources, as it actually \emph{increases} the storage required and substantially increases the computation required for training~\citep{li2018measuring}. \ours{} shares the idea of training only a subspace, but instead actively selects a productive subset of the dimensions of the original parameter space, making it possible to reduce both storage and compute requirements.

\subsection{Pruning}

Pruning networks to enforce sparsity after training has been effective in~\citet{handeep2015,zhusparsenn:2017, gecompressing2017, masanadomain-adaptive2017, luothinet:2017, ullrichsoft2017, lecun_optimal_1990, srivastavadropout:2014, wan_regularization_2013,EnergyAwarePruning:CVPR:2017}, while other work has focused on reducing the rank of the parameter matrices during training~\citep{alvarezcompression-aware2017} for later reduction. In contrast to \ours{}, pruning post-training requires a retraining step to regain lost accuracy, and both pruning and low rank constraints still require full dense backpropagation during the training phase, which is undesirable in embedded systems due to the extra memory and energy cost. \ours{} requires no retraining steps, and during both training and inference only ever stores the small subset of weights that are tracked. As a result, it only incurs a small portion of the total memory access costs during the training process.

Other work has focused on pruning while training to either improve accuracy or to increase the eventual sparsity of the trained network. \citet{zhu_prune_2017} gradually increase the number of weights masked from contributing to the network, while \citet{molchanov_variational_2017} extend variational dropout~\citep{kingma_variational_2015} with per-parameter dropout rates to increase sparsity. \citet{babaeizadeh_noiseout:_2016} inject random noise into a network to find and merge the most correlated neurons. Finally, \citet{langford_sparse_2008} decay weights every $k$ steps (for a somewhat large value of $k$), inducing sparsity gradually. Unlike \ours{}, all of these techniques require at least as much memory to train as the unpruned network initially, and all of them take longer to converge.

\subsection{Quantization}
Reducing storage costs is often approached through quantizing weights to smaller bit widths. This can be preformed either after training, as in~\citet{gecompressing2017, wuquantized2015, choitowards2016, handeep2015, ullrichsoft2017, gyselristretto:2016}, during an iterative retraining process~\citep{zhouincremental2017}, or during training, as in~\citet{caideep2017, zhoudorefa-net:2016, courbariauxbinarized2016, rastegarixnor-net:2016, gupta_deep_2015, mishra_wrpn:_2017, hubara_quantized_2016, courbariauxtraining2014, simard1994backpropagation, holt_back_1991}. Out of all of these methods, only~\citet{gupta_deep_2015, mishra_wrpn:_2017, hubara_quantized_2016, courbariauxtraining2014} use reduced precision \emph{while training} to lower the train time storage costs; other methods other methods store the full, non-quantized weights during backpropagation just like the post-training methods. Quantization is orthogonal to, and has been combined with, pruning techniques~\citep{handeep2015}; as such, it is also orthogonal to \ours{}.

\subsection{Other compression methods}
A few compression methods do not naturally fall into the quantization categories. HashedNets~\citep{DBLP:journals/corr/ChenWTWC15} use a hash function to group neuron connections into buckets with the same weight value. In effect, this is a kind of quantization, where the quantized values are limited in number but each is a full 32-bit floating-point number. Huffman coding~\citep{huffman_method_1952} has also been used for compression: for example, Deep Compression~\citep{handeep2015} applies Huffman coding to a pruned and quantized network. Both HashedNets and Deep Compression require the full network to be trained first, so are not directly comparable to \ours{}. Compression in general, however, is orthogonal to \ours{}.

\section{Conclusion}
Deploying DNN inference to mobile and other low-power devices has become possible in part due to advances in pruning and quantization of trained DNN models. The goal of \emph{training} on these devices has, however, remained elusive largely due to limitations on storage and bandwidth for both weights and activations.

In this paper, we focus on reducing the storage and bandwidth needed for weights. \ours{} is a training-time pruning technique that reduces weight storage both \emph{during} and \emph{after} training by (a)~tracking only the weights with the highest accumulated gradients, and (b)~recomputing the remaining weights on the fly.

Because \ours{} prunes exactly those weights that have learned the least, its weight diffusion profile during training is very close to that of standard (unconstrained) SGD, in contrast to other pruning techniques. This allows \ours{} to achieve better accuracy and weight reduction than prior methods on dense modern networks like Densenet, WRN, and ResNet, which have proven challenging to prune using existing techniques: \ours{} is able to reduce weight counts 5$\times$--11$\times$ with no accuracy loss.

Perhaps most attractively, \ours{} has the potential to dramatically reduce the memory footprint and memory bandwidth needed to store and access weights \emph{during} training. Because of this, \ours{} can potentially be used to train networks an order of magnitude larger than currently possible with typical hardware, or, equivalently, to train/retrain standard-size networks on limited-capacity hardware, something not possible with current training techniques.

\setlength{\bibsep}{0pt plus 0.3ex}

\bibliography{continuous-pruning-sysml2019}
\bibliographystyle{sysml2019}

\end{document}